%% file: llm_judge_relevance.tex
\documentclass[
]{ceurart}

\usepackage{listings} 
\usepackage{xcolor} 

\usepackage{tcolorbox} 

\usepackage{ulem} 

\usepackage{booktabs}
\usepackage{tabularx}

\begin{document}

\copyrightyear{2022}
\copyrightclause{Copyright for this paper by its authors.
  Use permitted under Creative Commons License Attribution 4.0
  International (CC BY 4.0).}

\conference{MMSR'24}

\title{Evaluating Cost-Accuracy Trade-offs in Multimodal Search Relevance Judgements}


\author{Silvia Terragni}[%
email=silvia@objective.inc
]

\author{Hoang Cuong}[%
email=hoang@objective.inc
]

\author{Joachim Daiber}[%
email=jo@objective.inc
]

\author{Pallavi Gudipati}[%
email=pallavi@objective.inc
]

\author{Pablo N. Mendes}[%
email=pablo@objective.inc,
]
\cormark[1]
\address{Objective, Inc. San Francisco, CA, 
USA.}

\cortext[1]{Corresponding author.}


\begin{abstract}
  Large Language Models (LLMs) have demonstrated potential as effective search relevance evaluators.
  However, there is a lack of comprehensive guidance on which models consistently perform optimally across various contexts or within specific use cases.
In this paper, we assess several LLMs and Multimodal Language Models (MLLMs) in terms of their alignment with human judgments across multiple multimodal search scenarios. Our analysis investigates the trade-offs between cost and accuracy, highlighting that model performance varies significantly depending on the context. Interestingly, in smaller models, the inclusion of a visual component may hinder performance rather than enhance it. These findings highlight the complexities involved in selecting the most appropriate model for practical applications.
\end{abstract}

\begin{keywords}
  Multimodal Search \sep
  Relevance Judgments \sep
  Large Language Models \sep
  Multimodal Large Language Models
\end{keywords}

\maketitle

\section{Introduction}
\input{1_introduction}

\section{Related Work}
\input{2_related}

\section{Methodology}
\input{3_methodology}

\section{Results}
\input{4_results}


\section{Conclusion}

\input{6_conclusion}




\begin{acknowledgments}
  Thanks to the entire Objective team for building many pieces of the puzzle that made this work possible. Special thanks to Lance Hasson, Brian Porter, George Gkotsis, Kuei-da Liao, and Faizaan Merchant. We would also like to thank Yev Rotar, John Gulley, and Brian Ip for their valuable relevance judgment inputs.
\end{acknowledgments}

\bibliography{llm_judge_relevance}

\appendix
\input{appendix}


\end{document}

%% file: 1_introduction.tex
Search relevance evaluation is the process of assessing how effectively an information retrieval system returns results that are relevant to a user's search query.
The process typically involves multiple human judges, tasked with stating whether each search result is relevant to a search query.
The resulting relevance judgments are then aggregated through evaluation metrics to quantify relevance.
Those in turn enable researchers and practitioners to compare different retrieval systems in order to select the best option for a given application.

Multimodal Search presents additional challenges in search relevance evaluations due to the complexity of interpreting and integrating information from various attributes across different modalities.
For instance, in e-commerce search, assessing relevance requires understanding the intent behind the search query and comparing it with a judge's interpretation of product relevance based on multiple features including the title, description, and images, as well as other attributes such as category, color, and price.

The task is further complicated by different characteristics across use cases.
For instance, when searching for very visual aspects (e.g. design assets) the images play a much more central role, as compared to other use cases where product category and other attributes are more important (e.g. searching for hotel supplies).
Data quality also varies significantly by use case. 
In applications with user-generated content, data may be missing or low quality -- e.g. product descriptions often conflict with the information that can be gleaned from images.

While human annotators remain the most reliable source for obtaining relevance judgments, the process is costly and time-consuming.

Recent work \cite{llm4eval}\cite{mllm-as-judge} has shown that Large Language Models (LLMs) and Multimodal Language Models (MLLMs) are viable alternatives for producing relevance judgments.
LLMs-as-judges are enticing since they can unlock higher relevance judgment throughput at a fraction of the cost.
As a result, they offer the potential of widespread relevance improvement in search systems due to more accessible and extensive evaluations, as well as training data generation.
However, progress is hampered by a number of under-explored questions about how to best employ LLMs-as-judges.

In this paper, we evaluate a number of LLMs and MLLMs in terms of their alignment with human judgments and ask the following research questions:

\begin{enumerate}
    \item Is LLM performance use-case dependent? In other words, would the same LLM perform well in one use case but not in another?
    \item Is there a clear winner? In other words, is there a model that consistently outperforms all the others across all use cases?
    \item Is multimodal support necessary for search relevance judgment in multimodal search?
    \item What models offer the optimal cost-accuracy trade-offs?
\end{enumerate}

In the next section we summarize related work. We then present our experimental setting and discuss results. Finally, we present concluding remarks and future work.

%% file: 2_related.tex
Large Language Models (LLMs) have shown exceptional abilities in a wide variety of tasks, and using them for evaluating Information Retrieval systems is receiving considerable attention \cite{llm4eval}. Recent studies have explored different methods for generating relevance judgments. For example, Prometheus \cite{prometheus} is a 13-billion parameter LLM designed to evaluate long texts using customized scoring rubrics provided by users. JudgeLM \cite{judgelm} uses fine-tuned LLMs as scalable judges to evaluate other LLMs effectively in open-ended tasks. They find that JudgeLM has a high agreement with expert judges, over 90\%, and works well in evaluating single answers, multimodal models, multiple answers, and multi-turn dialogues. Thomas et al. \cite{thomas2024llmsearchpref} developed an LLM prompt based on feedback from search engine users. They show accuracy similar to human judges and can identify difficult queries, best results, and effective groupings. They also find that both changes to prompts and simple paraphrases can improve accuracy.

In the context of Multimodal LLMs (MLLMs), Chen et al. \cite{mllm-as-judge} assess these models as judges through a new benchmark. They examine their performance in tasks such as Scoring Evaluation, Pair Comparison, and Batch Ranking. The study points out that MLLMs need more improvements and research before they can be fully trusted, as they can have biases like ego-centric bias, position bias, length bias, and hallucinations. Additionally, Yang et al. \cite{yang2024vlm} investigates the relevance estimation of Vision-Language Models (VLMs), including CLIP, LLaVA, and GPT-4V, within a large-scale ad hoc zero-shot retrieval task aimed at multimedia content creation.

To the extent of our knowledge, we are the first to compare the cost-accuracy trade-offs of several generally available LLMs of different sizes.

%% file: 3_methodology.tex
This study employs a relevance evaluation process to assess the performance of LLMs and MLLMs (collectively referred to as ``models") for search relevance judgments. 
We assess these models based on two critical dimensions: accuracy and costs. Our evaluation pipeline consists of three stages: 
\begin{itemize}
    \item \textit{Data Collection}: We obtained search results from three datasets across different domains using a list of predefined queries.
    \item \textit{Human Annotation}: Two trained human annotators assigned relevance grades to each (query, result) pair following some established relevance criteria.
    \item \textit{Model Evaluation}: We applied a range of LLMs and MLLMs to generate relevance judgments for the same sets of search results, comparing their performance against human annotations.
\end{itemize}
Each stage is discussed in detail in the following subsections, covering the datasets, retrieval system, grading strategy, and the models used.

\subsection{Datasets}
We conducted our experiments on three datasets: \textit{Fashion}, \textit{Hotel Supplies}, and \textit{Design}. The Fashion dataset is a subset of the publicly available dataset \textit{H\&M Personalized Fashion Recommendations}\footnote{\url{https://www.kaggle.com/competitions/h-and-m-personalized-fashion-recommendations}. For access to our human annotations for this dataset, please reach out to the corresponding author.}. The Hotel Supplies and Design datasets are proprietary and represent domains in the e-commerce search for hotel supply products, and social media search for design assets, respectively.
Each dataset includes multiple textual fields per product, along with one or more associated images. Table~\ref{tab:datasets} summarizes the characteristics of each dataset, detailing the average number of fields per search result, the average number of empty fields, and the average word count per result. These factors can impact the difficulty of generating relevance judgments.

\begin{table}[ht]
\centering
\begin{tabularx}{\textwidth}{l*{4}{>{\raggedleft\arraybackslash}X}}
\toprule
\textbf{Dataset} & \textbf{Total Number of Search Results} & \textbf{Avg Number of Textual Fields} & \textbf{Avg Number of Empty Textual Fields} & \textbf{Avg Number of Words per Result} \\ 
\midrule
Fashion & 1120 & 33 & 1 & 49 \\
Hotel Supplies  & 2210 & 17 & 8 & 96 \\
Design  & 1713 & 32 & 3 & 69 \\ 
\bottomrule
\end{tabularx}
\label{tab:datasets}
\caption{Summary statistics of the used datasets.}
\end{table}

\subsection{Retrieval System and Evaluation}
To obtain relevant search results, we utilized a baseline retrieval system that combines BM25 with BGE M3 embeddings ~\cite{bge2024}, which is one of the top-ranked text embedding models in the MTEB Leaderboard ~\cite{mteb} as of June 2024.
We created indexes for each dataset to enable efficient retrieval of results based on a predefined list of queries. These queries were either derived from real traffic data or carefully crafted by human experts to ensure they represented a wide range of search scenarios.
Our aim was to include queries and results that included hits and misses generated by both lexical and semantic retrievers.

\subsection{Relevance Judgment Strategy}

\begin{table}[ht]
\begin{tabular}{ccc}
\toprule
\textbf{Image} & \textbf{Search Result} & \textbf{Relevance Judgment} \\ \midrule
 \begin{tabular}[c]{@{}c@{}}
\includegraphics[width=0.2\textwidth]{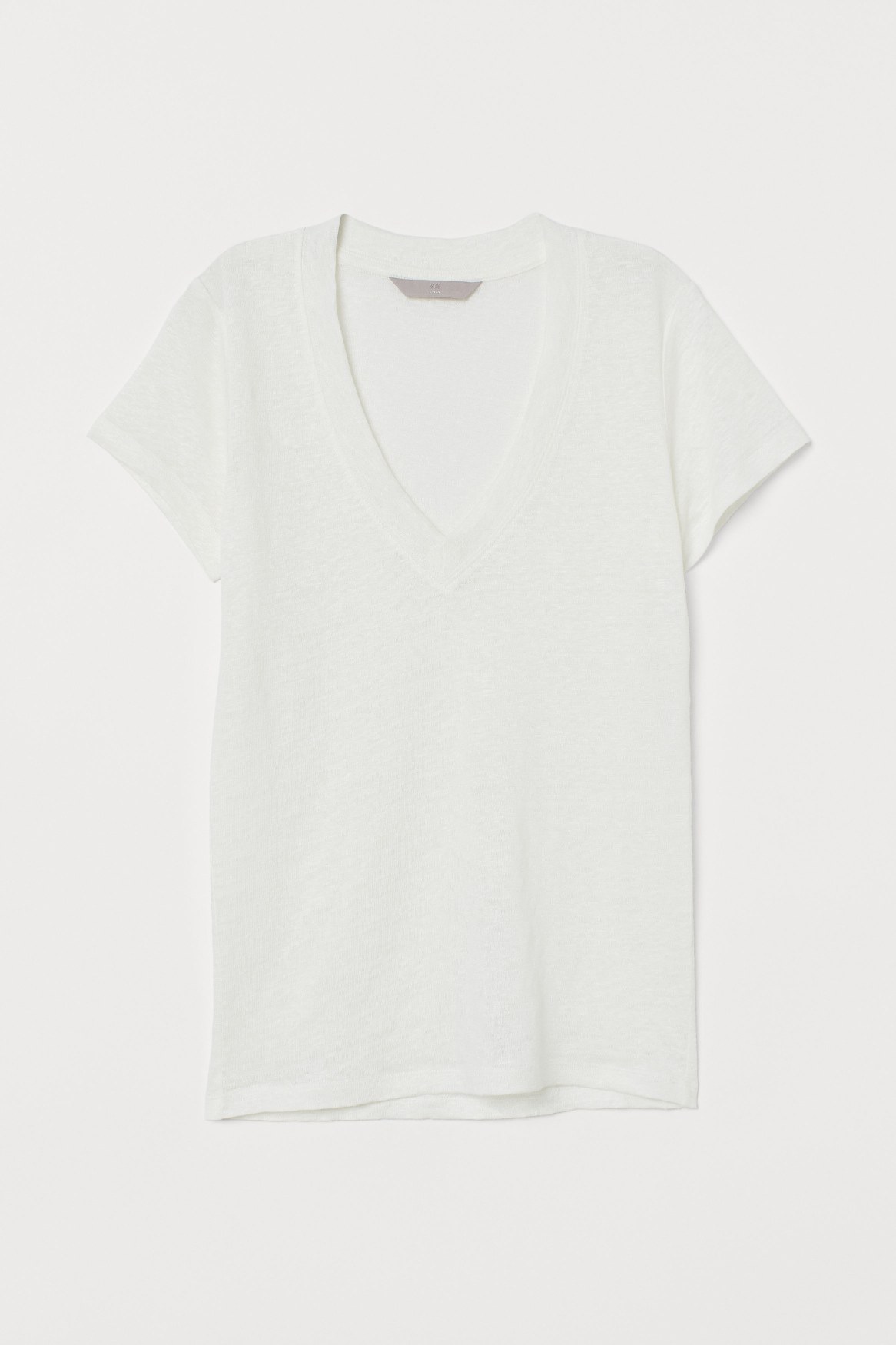} \end{tabular}&
  \begin{tabular}[l]{@{}l@{}}prod\_name: Premium ELKE vneck tee,\\    index\_name: Ladieswear,\\ detail\_desc: V-neck T-shirt in airy slub lin[...],\\ department\_name: Jersey/Knitwear Premium,\\     index\_group\_name: Ladieswear,\\  colour\_group\_name: White,\\   product\_type\_name: T-shirt,\\    graphical\_appearance\_name: Solid,\\  perceived\_colour\_value\_name: Light,\\   perceived\_colour\_master\_name: White\end{tabular} &
  2 \\ \midrule
 \begin{tabular}[c]{@{}c@{}}
\includegraphics[width=0.2\textwidth]{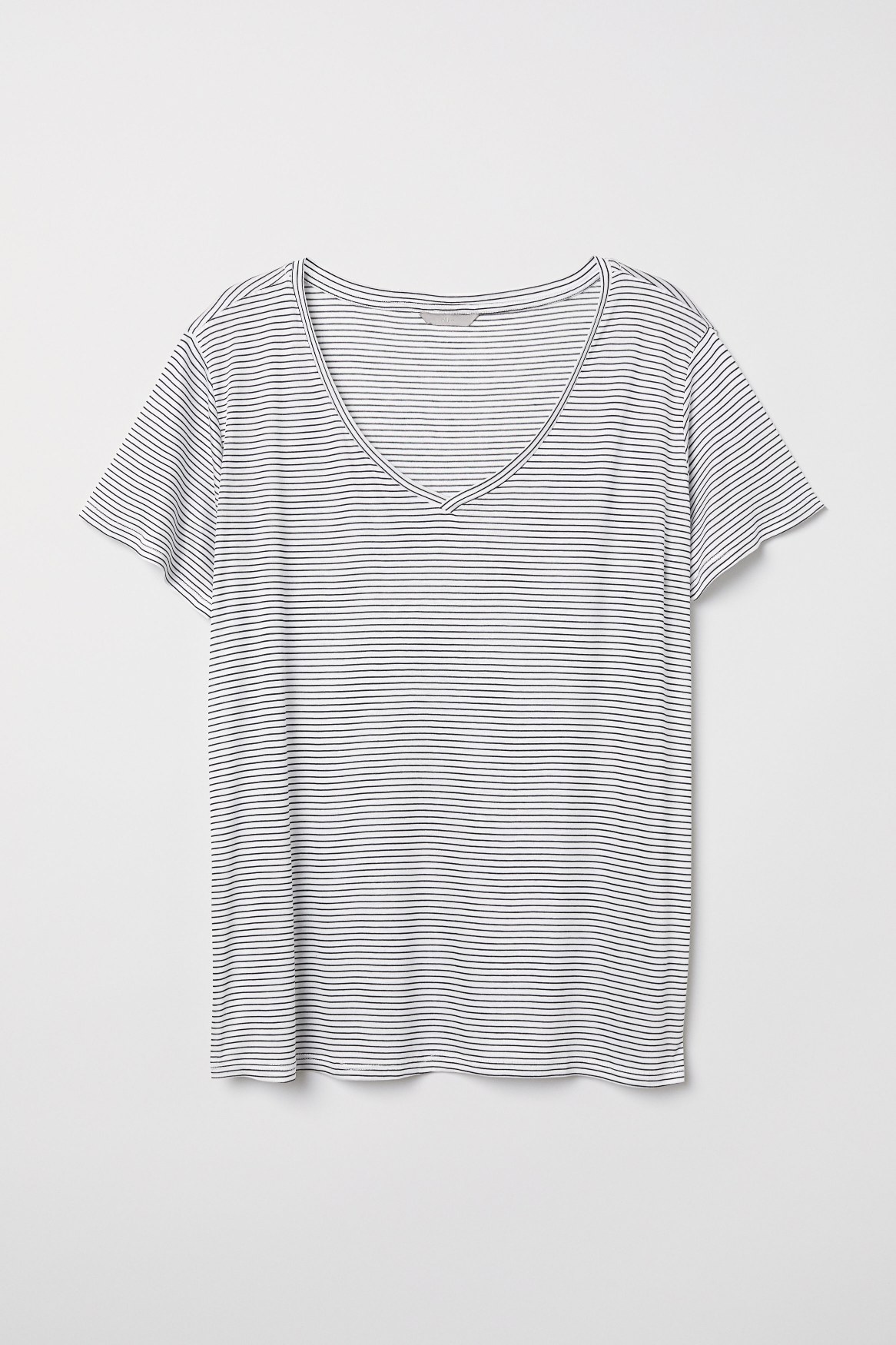}
\end{tabular}&
  \begin{tabular}[c]{@{}l@{}}prod\_name: ED Lizzie tee,\\  index\_name: Ladieswear,\\    detail\_desc: Short-sleeved top in lightwei[...],\\   department\_name: Jersey,\\   index\_group\_name: Ladieswear,\\  colour\_group\_name: White,\\   product\_type\_name: T-shirt,\\   graphical\_appearance\_name: Stripe,\\  perceived\_colour\_value\_name: Light,\\   perceived\_colour\_master\_name: White\end{tabular} &
  1 \\ \midrule
  \begin{tabular}[c]{@{}c@{}}
\includegraphics[width=0.2\textwidth]{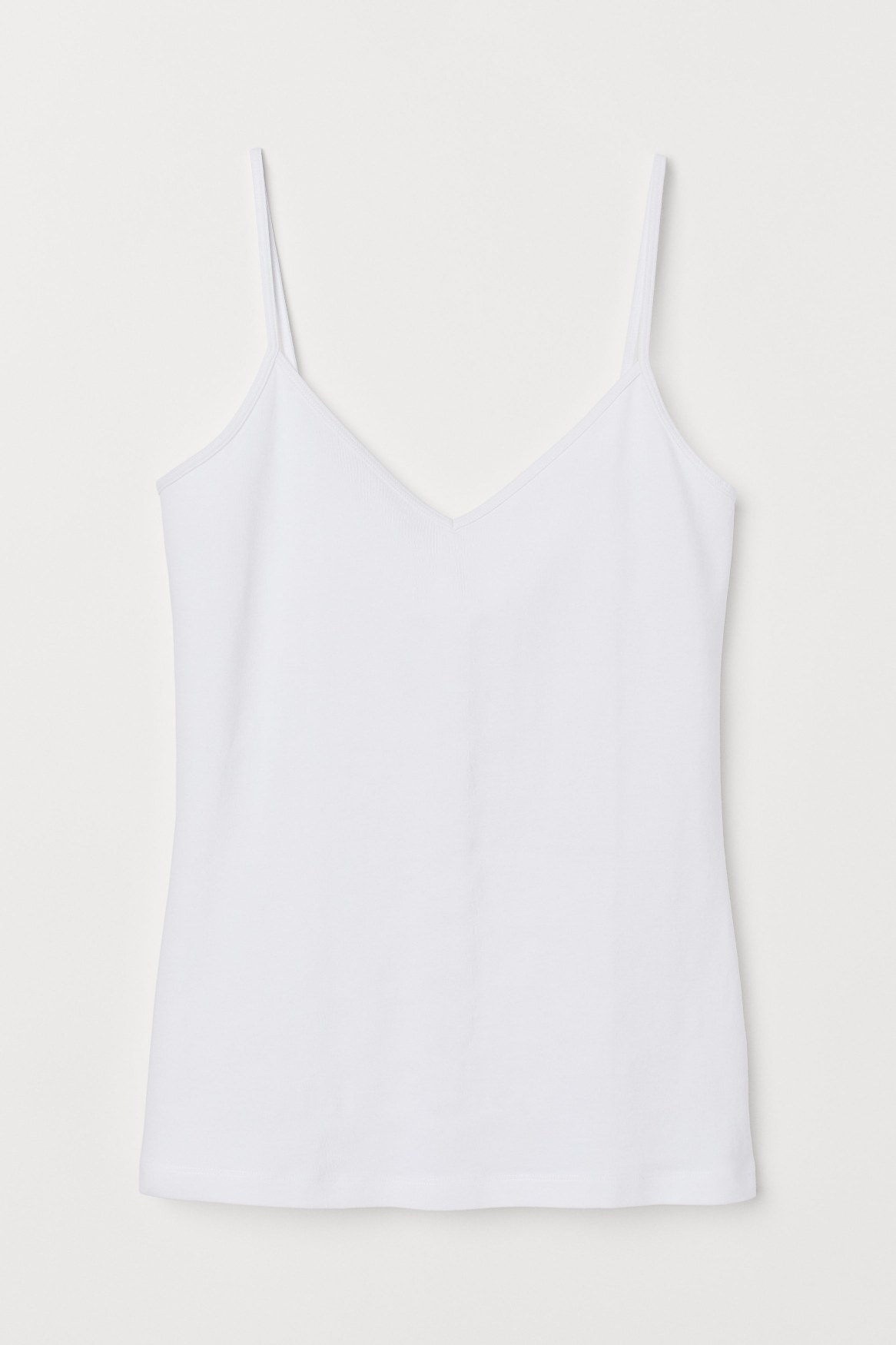} \end{tabular}&
  \begin{tabular}[l]{@{}l@{}}   prod\_name: V-neck Strap Top,\\   index\_name: Ladieswear,\\  detail\_desc: V-neck top in soft organic [...],\\ department\_name: Jersey Basic,\\   index\_group\_name: Ladieswear,\\ colour\_group\_name: White,\\   product\_type\_name: Vest top,\\  graphical\_appearance\_name: Solid,\\    perceived\_colour\_value\_name: Light,\\   perceived\_colour\_master\_name: White\end{tabular} &
  0 \\ \bottomrule
\end{tabular}
\label{tab:example_judgments}
\caption{Examples of three relevance judgment categories for the query ``\textit{v-neck white tee}'', accompanied by corresponding search results. The descriptions of the search results have been shortened for brevity.}
\end{table}

Each dataset was structured as a collection of query-result pairs. Two expert human annotators assessed the relevance of each pair on a 0-2 rating scale: 
\begin{itemize}
    \item 2: Highly relevant, a perfect match for the query
    \item 1: Somewhat relevant, a result that partially matches the query’s intent
    \item 0: Not relevant, a poor result that should not be shown
\end{itemize}
The human annotators were provided with detailed guidelines to ensure consistency in their relevance judgments. Table~\ref{tab:example_judgments} provides examples of different relevance judgment categories for the query ``\textit{v-neck white tee}''. In the first row, the result is highly relevant, as both the image and the text describe a white v-neck t-shirt. Therefore the human relevance judgment for this pair is a 2. The second row shows a partial match: while the text mentions a white t-shirt, the image depicts a white v-neck t-shirt with black stripes, resulting in a relevance judgment of 1. The third row illustrates an irrelevant result (0), where the product shown is a strap top, unrelated to the query.

In the Appendix, we describe in detail our internal H\&M grading guidelines that human annotators follow to assign relevance grades to query-result pairs. We note that our grading guidelines are based on similar principles to those used for the Hotel Supplies and Design datasets. However, the guidelines have been also adapted to suit the specific characteristics of the datasets.

\subsection{Inter-Annotator Agreement}
To assess the reliability of the relevance judgments, either human or LLM-generated, we followed common practice and calculated \textit{Cohen’s kappa} coefficient. Cohen's kappa is a robust statistical measure commonly employed to quantify inter-annotator agreement for categorical data. Cohen's kappa values range from -1 to 1, where 1 indicates strong agreement, while values closer to 0 suggest agreement no better than chance. To interpret the kappa values, we use the guidelines reported in Table~\ref{tab:cohenkappa}.

\begin{table}[h]
\begin{tabular}{@{}ll@{}}
\toprule
\textbf{Cohen’s kappa} & \textbf{Interpretation}  \\ \midrule
0 - 0.20               & Slight agreement         \\
0.21 - 0.40            & Fair agreement           \\
0.41 - 0.60            & Moderate agreement       \\
0.61 - 0.80            & Substantial agreement    \\
0.81 - 1.00            & Almost perfect agreement \\ \bottomrule
\end{tabular}
\caption{Guidelines for interpreting Cohen's kappa values.}
\label{tab:cohenkappa}
\end{table}

In our evaluation, we compute Cohen’s kappa to measure the agreement between human annotators and LLM-generated annotations, as well as between pairs of human annotators. The degree of agreement between human annotators also provided insights into the difficulty of evaluating certain datasets.

\subsection{Models}
Our evaluation included a range of LLMs and MLLMs to reflect varying levels of performance and cost. We considered both large-scale proprietary models and more cost-efficient alternatives:
\begin{itemize}
    \item OpenAI Models\footnote{\url{https://platform.openai.com/docs/models}}: {GPT-4V (\textit{gpt-4-vision-preview}), GPT-4o (\textit{gpt-4o-2024-05-13}), GTP-4o-mini (\textit{gpt-4o-mini-2024-07-18})}
    \item Anthropic Models\footnote{\url{https://docs.anthropic.com/en/docs/about-claude/models}}: Claude 3.5 Sonnet, Claude 3 Haiku
\end{itemize}

\subsection{Prompts}

To design the prompts for the models under consideration, we created a template aimed at guiding the models to generate accurate relevance judgments. 

In the multimodal setup, where an image is provided, the prompt will reference and include the image.
Additionally, we require the model to provide an explanation for its relevance judgment. This element could be useful for interpreting the model's decisions. 

Below, we present the prompt used for the Claude family in the text-only scenario. For the complete set of prompts, please refer to the Appendix. In the template, \textit{\{\{document\}\}} and \textit{\{\{query\}\}} are placeholders for the search result and query, respectively.
\begin{tcolorbox}[colback=gray!5!white, colframe=gray!75!black, title=Haiku's Prompt Template (Text-only Setup)]
\begin{lstlisting}[basicstyle=\ttfamily, frame=none]
You are an assistant responsible for rating how the retrieved result is relevant to
the query. Output a token: "2", "1", or "0" followed by a full explanation.
Guidelines:
"2" - The result matches exactly with what the user's query is looking for.
"1" - The result is not exactly with what the user's query is looking for. But it's
pretty similar. As our aim is to be strict on exact matches, this grade is less
likely to be used.
"0" - The result is not related to the query at all.

Result: {{document}}
Query: {{query}}
Output:"
\end{lstlisting}
\end{tcolorbox}

It is important to note that these prompt templates are not the result of an extensive exploration of all possible templates.
In Section~\ref{sec:prompt_engineering}, we provide an analysis of the prompt engineering process that led to the best-performing prompts for Haiku.

%% file: 4_results.tex
\subsection{Multimodal vs Single-modality Evaluation}

\begin{table}[ht]
\begin{tabular}{@{}lrrrrrrrrrrc@{}}
\toprule
\textbf{} & \multicolumn{2}{c}{\textbf{GPT-4v}} &
  \multicolumn{2}{c}{\textbf{GPT-4o}} &
  \multicolumn{2}{c}{\textbf{GPT-4o-mini}} &
  \multicolumn{2}{c}{\textbf{Sonnet}} &
  \multicolumn{2}{c}{\textbf{Haiku}} &
  \multicolumn{1}{l}{\textbf{Human}} \\ \midrule
\textbf{} & \multicolumn{1}{l}{MM} &
  \multicolumn{1}{l}{Text} &
  \multicolumn{1}{l}{MM} &
  \multicolumn{1}{l}{Text} &
  \multicolumn{1}{l}{MM} &
  \multicolumn{1}{l}{Text} &
  \multicolumn{1}{l}{MM} &
  \multicolumn{1}{l}{Text} &
  MM &
  \multicolumn{1}{l}{Text} & MM \\  \midrule
Fashion & 0.503 &
  0.498 &
  \textbf{0.613} &
  0.606 &
  0.424 &
  0.382 &
  0.441 &
  0.387 &
  0.371 &
  0.431 &
  0.680\\
  Hotel Supplies &
0.620 &
  0.596 &
  0.627 &
  0.582 &
  0.506 &
  0.565 &
  0.634 &
  \textbf{0.638}&
  0.471 &
  0.560 &
  0.641 \\
Design & 
0.320 &
  0.317 &
  \textbf{0.404} &
  0.331 &
  0.294 &
  0.299 &
  0.351 &
  0.381 &
 0.260 &
  0.309 &
  0.447
  \\  \midrule
Average &
0.481 &
  0.471 &
  \textbf{0.548} &
  0.506 &
  0.408 &
  0.415 &
  0.475 &
  0.469 &
  0.368 &
  0.433 &
  0.589 \\ \bottomrule
\end{tabular}
\caption{Cohen's kappa coefficients between one of the human annotators and the considered Multimodal (MM) models and their text-only (Text) counterparts. The last column shows the inter-annotator agreement between the two human annotators. }
\label{tab:mm_vs_text_correlation}
\end{table}
The results presented in Table~\ref{tab:mm_vs_text_correlation} offer several insights into the performance of the considered Large Language Models across different domains and modalities. 

\paragraph{Use-case Dependency of LLM Performance} 
The analysis reveals that the LLM performance is dependent on the use case. The models show varying levels of correlation with the human relevance judgments across the different domains. For example, GPT-4V shows higher performance in the Hotel Supplies use case but performs relatively worse in the other areas. We can observe a similar trend across the other models. This variability in model performance is also connected to the inherent difficulty of the tasks. This is also reflected by the varying levels of agreement among the human annotators for the different use cases. 

\paragraph{One Model to Rule Them All? }
The multimodal version of GPT4-o generally performs better than the other models in two out of three cases, achieving the highest average Cohen's kappa coefficient (0.548). It stands out in the Hotel Supplies and Fashion domains, where it shows substantial agreement with human annotations. 
However, it is outperformed by Sonnet in the Hotel Supplies domain, suggesting that no single model outperforms all the others across every use case.  

\paragraph{Tailor Model Prompt to a Specific Dataset}
{We observe that tailoring a model's prompt to a specific domain can greatly improve its grading performance on the corresponding dataset. A notable example is the text-only Haiku LLM. Despite it being among the least powerful models in our experiments, we achieved the highest Cohen’s kappa coefficient for that dataset ($0.6403$ compared to $0.560$) by refining the prompt for the Hotel Supplies dataset. Nevertheless, we also note that using the Hotel Supplies dataset-adjusted prompt may lead to significant overfitting to other datasets. For example, when using the same prompt for the Design dataset, Cohen's kappa coefficient decreases to $0.333$ (instead of $0.431$).}

\paragraph{Necessity of Multimodal Support}
In the table, we compare each Multimodal (MM) model with its text-only (Text) counterpart. It is worth noticing that the benefits of multimodal support are not uniform across all the models and use cases. For models like GPT-4o, the vision component significantly enhances the performance, increasing the correlation from 0.506 (Text) to 0.548 (MM). This leads to the highest average performance and the metric is remarkably very close to human correlation (i.e. 0.589). GPT-4V and Sonnet also benefit from the visual component. However, for smaller models, such as Haiku, the vision component appears to have a detrimental effect, decreasing the correlation from 0.433 (Text) to 0.368 (MM). To further investigate the impact of the visual component in Haiku, we performed an ablation study by excluding the textual component and relying solely on the image. Under this configuration, the highest correlation achieved was 0.1 for the Design case, which is significantly lower than the text-only correlation of 0.309. This indicates that for smaller models like Haiku, the visual component may not be sufficiently robust to provide effective multimodal support.

\paragraph{Error Analysis}
To investigate the poor performance of the smaller multimodal models, we conducted an error analysis on a sample of relevance judgments from Haiku that disagreed with both human annotators. We examined 31 instances of disagreement and identified three distinct error categories. Notably, Haiku generates an explanation for its relevance judgments, which allows us to categorize the errors effectively.

The most frequent issue (17 cases) involved the model’s failure to correctly identify the product type. For example, when given the query ``\textit{pure cotton dressing gown}'', the model misclassifies a linen dressing gown, justifying its choice with the explanation: \textit{The product is [...] made of linen, which is a natural fiber similar to cotton}. In half of the remaining cases, Haiku's errors originate from wrong assumptions. For instance, the model confused the brand name for bras with the word ``band'' as in ``hairband'', leading to incorrect judgments. Lastly, 7 of the cases were related to the model’s vision capabilities, where it failed to recognize specific patterns or prints on products, resulting in inaccurate relevance assessments. Table~\ref{tab:example_wrong_judgement} provides an example of this type of error, including the image, query, and explanation generated by Haiku.

\begin{table}[h]
\centering
\begin{tabular}{llp{5.3cm}}
\toprule
Image &
  Query &
  Explanation \\ \midrule
  \begin{tabular}[c]{@{}c@{}}
\includegraphics[width=0.2\textwidth]{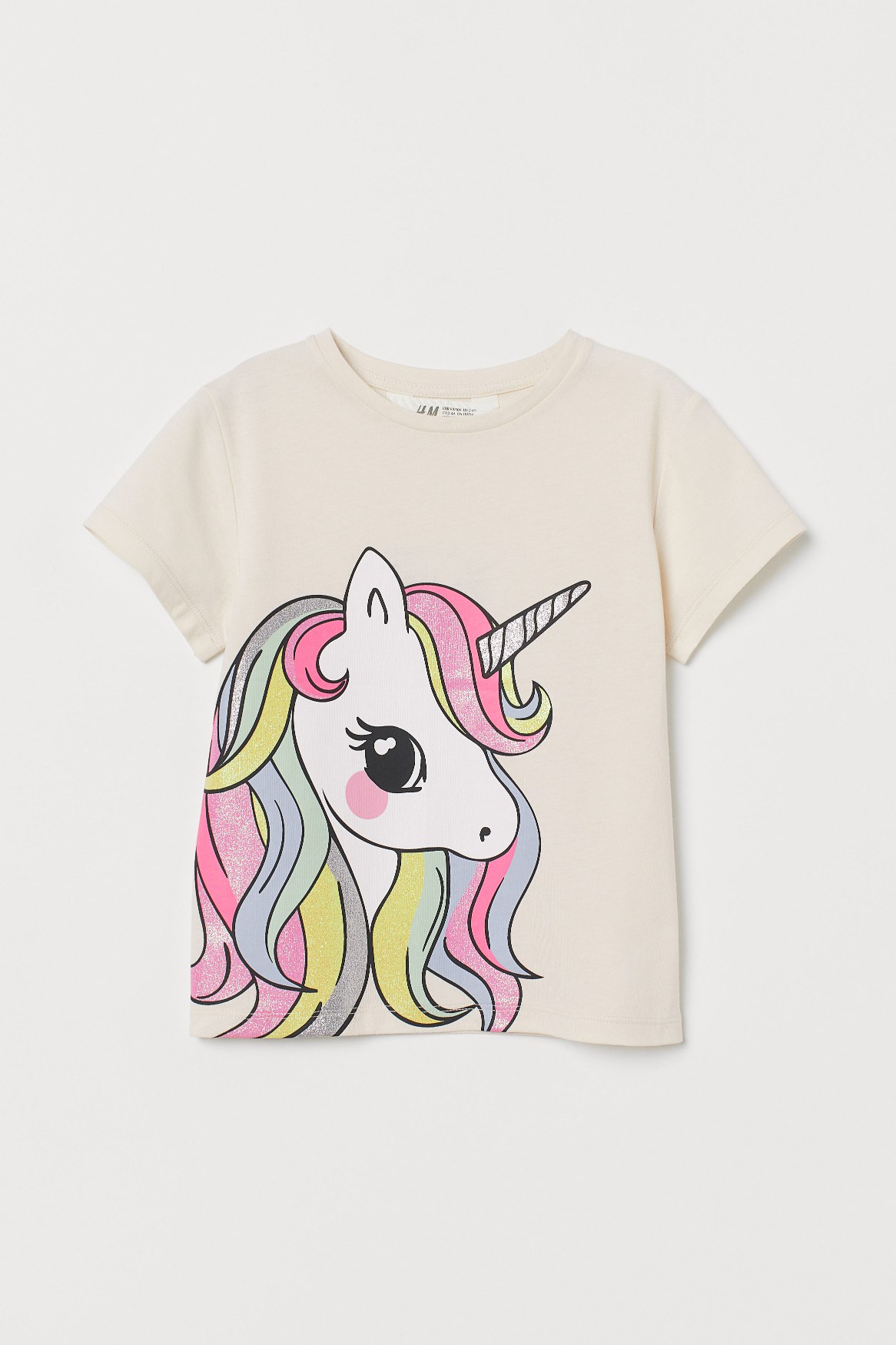} \end{tabular} &
  h\&m kids unicorn printed t-shirt &
   \begin{tabular}[c]{@{}p{5.3cm}@{}}
  The provided result does not match the user's query for a ``h\&m kids unicorn printed t-shirt''. The result is for a ``Mia l/s top'' which is a long-sleeved top in soft, printed cotton jersey, but it does not appear to have a unicorn print. \textbf{The image also does not show a unicorn print.} Therefore, the result is not relevant to the user's query, and I would rate it as a ``0''. 
  \end{tabular} \\ \bottomrule
\end{tabular}
\caption{Example of an error in Haiku’s relevance judgment due to its inability to understand the provided image.}
\label{tab:example_wrong_judgement}
\end{table}

\paragraph{Cost-Accuracy Trade-off}

\begin{table}[ht]
\begin{tabular}{@{}lrrrrr@{}}
\toprule
                              & GPT-4V & GPT-4o & GPT-4o-mini & Sonnet  & Haiku \\ \midrule
\$/1M Input tokens            & 10.00  & 5.00   & 0.15        & 3.00    & 0.25  \\
\$/1M Output tokens           & 30.00  & 15.00  & 0.60        & 15.00   & 1.25  \\
\$/1M images (low resolution) & 425.00 & 425.00 & 425.00      & 1048.58 & 87.38 \\ \bottomrule
\end{tabular}
\caption{Cost comparison across different models for input tokens, output tokens, and image processing. The costs are reported per million units, with image processing based on low-resolution images. Prices reflect the providers' rates as of August 16, 2024.}
\label{tab:costs}
\end{table}
Considering the previous results comparing multimodal versus text-only performance, we can make important cost-accuracy trade-off considerations when choosing a model to adopt for relevance judgment.
The costs reported in Table \ref{tab:costs} reflect the providers' pricing as of August 16, 2024. For image processing, calculations are based on handling 1M low-resolution images. {Specifically, OpenAI's GPT-4V and GPT-4o allow users to use a low-resolution with 512x512 pixels of the image and represent it with a budget of 85 tokens. This results in a cost of \$0.000425 per image.\footnote{For GPT-4o-mini, this limit is set at 2,833 tokens (instead of 85 tokens) per image and this leads to the same per-image cost.} For fairness to Claude models, we thus also report their prices for images resized to 512x512 pixels.}

In this setup, the costs for image processing are fixed per search result, while the input and output tokens are variable, depending on the length of the search result being evaluated. This means that, when evaluating 1M images, we have a fixed cost of \$425 for the GPT family, approximately \$1048 for Sonnet, and \$87 for Haiku. To these fixed costs, we must add variable expenses, depending on the prompt, the search result, and output lengths. As Table~\ref{tab:datasets} shows, our datasets contain search results with varying word counts. Additionally, different models use varying prompts and different tokenizers, leading to differences in the number of tokens.

Along with its strong performance for both text and multimodal tasks, GPT-4V is the most expensive model with high costs for tokens and image processing. With an average of 867 input tokens, for 1M multimodal search results, the cost for processing 1M multimodal search results with GPT-4V is \$425 (image cost) + $867 \cdot \$10.00$ (input token cost) = \$9,095.

In contrast, GPT-4o offers higher performance at a lower cost. {With an average of 889 input tokens per result, the cost for processing 1M multimodal results is approximately \$425 (image cost) + $889 \cdot \$5$ (input token cost) = \$4,865.} This significantly lower cost (i.e. 50\% of the cost of using GPT-4V) makes it the current best choice when high precision is required. 

For Sonnet, with an average of 784 input tokens, the cost for processing 1M multimodal results would be \$1048.58 (image cost) + $784 \cdot \$3.00$ (input token cost) = \$3400.58. Given Sonnet's performance as the third-best model in terms of correlation with human evaluations, it represents a suitable choice for scenarios where a moderate budget is available but maintaining high-quality results is still important.

For smaller models like GPT-4o-mini and Haiku, the cost differences become significant, though at the expense of performance. In a text-only setting, {GPT-4o-mini offers the lowest cost per result, making it an attractive option for large-scale applications where lower accuracy can be tolerated. While Haiku’s cost per result is slightly higher than GPT-4o-mini, its performance is also superior. However, our experiments indicate that the visual component did not significantly enhance the performance of these smaller models for relevance judgment. In fact it can be even detrimental when using the visual component. Therefore, the multimodal capabilities of GPT-4o-mini and Haiku should be employed with caution,} especially considering the high costs associated with image processing—particularly for GPT-4o-mini. 

\subsection{Prompt Engineering}\label{sec:prompt_engineering}

We made the following observations:

\paragraph{Strictness Guidelines} Many of the initial disagreements with humans stemmed from the models being more lenient about the \textit{1} (OK) grade. Results improved after we appended instructions to prefer grades \textit{2} (GREAT) and \textit{0} (BAD) -- e.g. ``As our aim is to be strict on exact matches, this grade is less likely to be used."

\paragraph{Smaller Models Are More Sensitive to Prompt Complexity} {We found that smaller models, such as Haiku, are highly sensitive to prompt complexity, whereas larger models like GPT-4V manage these complexities more effectively. For example, when using a prompt with comprehensive and somewhat lengthy grading instructions, we observed a significantly higher Cohen’s kappa coefficient for the Hotel Supplies dataset with GPT-4V (0.54) compared to Haiku (0.32).

Note that this does not imply that Haiku is not suitable for grading the task; rather, it suggests that the model performs better when prompts are simpler. We verify this hypothesis further by making prompts progressively more concise while still retaining the essential instructions. Ultimately, we were able to refine the prompt for the Hotel Supplies dataset that helps the text-only Haiku model achieve the highest Cohen’s kappa coefficient of $0.64$.}

\paragraph{Different Models May Work with Different Prompts}
{Although Haiku achieves the highest Cohen’s kappa coefficient of $0.64$ on the Hotel Supplies domain with the refined prompt we developed, we did not observe the same improvement with GPT-4V. When using the same prompt, GPT-4V maintained a similar Cohen’s kappa coefficient of approximately $0.54$. This indicates that prompt engineering can be highly model-specific, and a prompt that works exceptionally well for one LLM model may not perform as well for others. In fact, we could not find a systematic way to reliably optimize model accuracy across the board. As a result, the process of prompt engineering feels more like art than science and motivates further work to develop systematic ways to discover the upper limits of accuracy for each model size.}

\paragraph{Asking for explanations} {In our experiments, we observed that asking LLMs to provide explanations for grading outputs is beneficial for several important reasons:
\begin{itemize}
    \item Relevance is subjective, and asking an LLM to explain its grading output can be helpful in verifying the correctness of the assessment. In our experiments, it was not uncommon for humans to initially disagree with the grading outputs from LLMs; however, they often reached a consensus after reviewing the detailed explanations provided.
    \item Having an explanation also helps us to understand how to iterate via prompt engineering to make the instructions less ambiguous for the model.
\end{itemize}

We also note that asking LLMs to provide explanations may help the model perform better at grading. However, note that prompts need to be carefully crafted, as we also observed the performance may regress if we do not do it well.

In the end, we were able to meaningfully improve the accuracy of Haiku through prompt engineering that requests LLMs for explanations (from $0.36$ to $0.40$).} Given that this is not far in accuracy, and Haiku is $20$-$40$ times cheaper than the GPT family, this makes it a very appealing option for application at a large scale. For instance, smaller models can be used to generate larger label sets to explore recall issues, while more expensive models focus on smaller sets to evaluate precision.

%% file: 6_conclusion.tex
In this paper, we have presented a new analysis of MLLMs-as-a-Judge, to assess the cost-accuracy trade-offs of relevance judgment capabilities of MLLMs across three multimodal search use cases: Hotel Supplies, Design, and Fashion.
Various LLMs have shown potential, but no single LLM showed optimal cost-accuracy trade-off across all use cases evaluated.

We have found that for any given practitioner looking to choose the best LLM judge for their use case, a comprehensive evaluation of all available models is both time-intensive, financially demanding, and requires significant amounts of energy, which can have a significant effect on the environment.
This motivates future work in the following directions:
1) improving the abilities of general MLLMs across use cases,
2) improving cost and computational efficiency of large MLLMs, and
3) creating small MLLMs that are optimized for judging relevance in cost-optimal ways for more specialized applications.

%% file: appendix.tex
\section{Prompt Templates}
In this section, we report the prompts used for the considered models. 
In the templates, \textit{\{\{document\}\}}, \textit{\{\{query\}\}}, and \textit{\{\{image\}\}} are placeholders for the search result, query, and image respectively. For the OpenAI's models, the image corresponds to the image URL, while for the Anthropic's models, it corresponds to a base64-encoded image.



\begin{tcolorbox}[colback=gray!5!white, colframe=blue!75!black, title=Haiku and Sonnet's Prompt Template (Multimodal Setup)]

\begin{tcolorbox}[colback=gray!5!white, colframe=gray!75!black]
\begin{lstlisting}[basicstyle=\scriptsize\ttfamily, frame=none]
You are an assistant responsible for rating how the retrieved result is relevant to the query. If an image 
is available, use it to determine the relevance to the query. Output a token: "2", "1", or "0" followed by
a full explanation.
Guidelines:
"2" - The result matches exactly with what the user's query is looking for.
"1" - The result is not exactly with what the user's query is looking for. But it's pretty similar. As our
aim is to be strict on exact matches, this grade is less likely to be used.
"0" - The result is not related to the query at all.

Result: {{document}}
Query: {{query}}
\end{lstlisting}
\end{tcolorbox}

\begin{tcolorbox}[colback=gray!5!white, colframe=gray!75!black]
\begin{lstlisting}[basicstyle=\scriptsize\ttfamily, frame=none]
{{image}}
\end{lstlisting}
\end{tcolorbox}

\begin{tcolorbox}[colback=gray!5!white, colframe=gray!75!black]
\begin{lstlisting}[basicstyle=\scriptsize\ttfamily, frame=none]
Token:
\end{lstlisting}
\end{tcolorbox}
\end{tcolorbox}

\begin{tcolorbox}[colback=gray!5!white, colframe=blue!75!black, title=GPT4's Prompt Template (Multimodal Setup)]

\begin{tcolorbox}[colback=gray!5!white, colframe=gray!75!black, title=User Role: System]
\begin{lstlisting}[basicstyle=\scriptsize\ttfamily, frame=none]
You are a helpful assistant designed to output JSON. You are RateGPT, an intelligent assistant that can 
score search results based on their relevance to a query and the user's intent behind the query. You should
return JSON with two required fields 'reasoning' and 'score'. In the 'reasoning' field, you can explain
your observations of relevance. When producing a score, use the following grading criteria:
    - 0 (BAD) - Use this grade for a search result if it is not related to user's query at all.
    - 1 (OK) - This grade is for a search result that is not exactly what the user's query is looking for,
but it's pretty similar. As our aim is to be strict on exact matches, this grade is less likely to be used.
    - 2 (GOOD) - The product matches exactly with the user's intent and query. Use this score this if the
search result aligns perfectly with the user's query.
    
### Query Analysis
Before you start grading, it's essential to understand user's intent by breaking apart the query. Keep in
mind, some queries may be more explicit than others. For instance, if user is searching for a clothing
product, then "Red checkered jacket" is more specific compared to "Red jacket". Another example, if user is
searching for a venue, then "Rock concert in San Francisco this weekend" is more specific compared to "Rock
concert in San Francisco". Therefore, adapt your grading contextually.

Consider all the information from all fields. 

Note: All fields should be taken with equal importance. You should adhere strictly to these guidelines
while grading and ensure a holistic evaluation of
the search results based on all considered fields.
\end{lstlisting}
\end{tcolorbox}

\begin{tcolorbox}[colback=gray!5!white, colframe=gray!75!black, title=User Role: User]
\begin{lstlisting}[basicstyle=\scriptsize\ttfamily, frame=none]
You are given a search query and a search result in json format. If an image is available, use it to
determine the relevance to the query. You must indicate with a score whether the result is relevant or not.
---
Follow the following format. 

Query: {{query}}

Result: {{result}}

Score: 0 or 1 or 2

---

Query: {{query}}

Result: {{document}}
\end{lstlisting}
\end{tcolorbox}

\begin{tcolorbox}[colback=gray!5!white, colframe=gray!75!black, title=User Role: User]
\begin{lstlisting}[basicstyle=\scriptsize\ttfamily, frame=none]
{{image}}\end{lstlisting}
\end{tcolorbox}
\end{tcolorbox}

\section{H\&M grading guidelines}
{In this section, we introduce the internal H\&M grading guidelines that our human annotators follow to assign relevance grades to each pair of <query, search result>.

\subsection{Grading rules}
\begin{itemize}
    \item Is the query asking for a specific category of product, either narrow or broad? 
    \begin{itemize}
    \item If the answer is \textit{yes}, then the user is asking for results that are \textit{restricted} to that category. Only results matching that product category will be marked as \textit{GREAT}. If the results don't match the category, mark them as \textit{BAD}.
        \item It is the grader's job to identify the category as introduced within the query and classify products accordingly.
        \begin{itemize}
            \item For instance, examples of queries->categories can be \textit{``office clothes''}->\textit{``clothes''} or \textit{``music t-shirt''}->\textit{``t-shirt''}. Note that \textit{t-shirt} can be categorized as a sub-category of \textit{clothes} and clothes is a super-category of \textit{t-shirts}. As such, \textit{``office clothes''} includes more products since it implies a broader categorization, whereas \textit{``music t-shirt''} is restricting products under \textit{``t-shirts''}.
        \end{itemize}
    \end{itemize}
    \item Is the query mentioning a feature, for instance: \textit{color}, \textit{size}, \textit{utility} (e.g. \textit{windproof}, \textit{maternity})?
    \begin{itemize}
        \item If the query mentions a feature, only results matching that feature will be marked as \textit{GREAT}.
        \item If the results don't match the feature, mark it as \textit{BAD}.
        \item If the results match the feature close enough but not exactly, mark it as \textit{OK}. For example, if the query is \textit{``yellow jacket''} and a search result is a light orange jacket, or some jacket that contains some clear patches of yellow but is otherwise not yellow, then this result is \textit{OK}.
    \end{itemize}
    \item If a query mentions both a category and a feature, only results matching both the category and the feature will be marked as \textit{GREAT}. Results matching the feature but not the category (or vice-versa) are \textit{BAD}.
\end{itemize}
\subsection{Grading examples}
\begin{table}[h]
\centering
\begin{tabular}{llll}
\multirow{2}{*}{Query}
 &
 \multicolumn{3}{c}{Grading}\\
&  BAD &
  OK &
  GREAT \\\hline
  \textit{yellow raincoat} & \begin{tabular}{@{}l@{}}black raincoat yellow\\cardigan yellow jacket\end{tabular} &
  \begin{tabular}{@{}l@{}}\begin{tabular}{@{}l@{}}raincoat that is of different\\ but close enough color\\ (e.g. orange, or some other\\ color but clearly also\\ has patches of yellow)\end{tabular}\end{tabular} &\begin{tabular}{@{}l@{}}raincoats that are yellow\end{tabular}\\\hline

  \textit{rainy weather} & \begin{tabular}{@{}l@{}}anything else/products\\ that do not help during\\ a rainy weather\end{tabular} &
  \begin{tabular}{@{}l@{}}\begin{tabular}{@{}l@{}}a product that is not\\ saying explicitly it’s\\ rainproof or for rainy\\ weather but can be used to\\ protect from rain\end{tabular}\end{tabular} &\begin{tabular}{@{}l@{}}any product that is rainproof\end{tabular}\\\hline

  \textit{sports utility
wear} & \begin{tabular}{@{}l@{}}anything that cannot\\ be used during a sport\\ activity or some kind of\\ manual labour task.\end{tabular} &
  \begin{tabular}{@{}l@{}}\begin{tabular}{@{}l@{}}products that are\\ supporting a sports\\ lifestyle but do not\\ necessarily advertise\\ themselves as such \\(e.g. T-shirt)\end{tabular}\end{tabular} &\begin{tabular}{@{}l@{}}products that visually\\ and textually support\\ comfortable movement\\ and gym workouts\end{tabular}
\end{tabular}
\end{table}
}